\documentclass[conference]{IEEEtran}
\IEEEoverridecommandlockouts
\usepackage{cite}
\usepackage{amsmath,amssymb,amsfonts}
\usepackage{algorithmic}
\usepackage{graphicx}
\usepackage{textcomp}
\usepackage{xcolor}

\usepackage{epsfig}
\usepackage{multirow}

\def\BibTeX{{\rm B\kern-.05em{\sc i\kern-.025em b}\kern-.08em
    T\kern-.1667em\lower.7ex\hbox{E}\kern-.125emX}}
\begin{document}

\title{Unsupervised Domain Adaptation for Neuron Membrane Segmentation based on Structural Features\\
}

\author{\IEEEauthorblockN{Yuxiang An}
\IEEEauthorblockA{\textit{School of Computer Science} \\
\textit{The University of Sydney}\\
yuan5699@uni.sydney.edu.au}
\and
\IEEEauthorblockN{Dongnan Liu}
\IEEEauthorblockA{\textit{School of Computer Science} \\
\textit{The University of Sydney}\\
dongnan.liu@sydney.edu.au}
\and
\IEEEauthorblockN{Weidong Cai}
\IEEEauthorblockA{\textit{School of Computer Science} \\
\textit{The University of Sydney}\\
tom.cai@sydney.edu.au}
}

\maketitle

\begin{abstract}
AI-enhanced segmentation of neuronal boundaries in electron microscopy (EM) images is crucial for automatic and accurate neuroinformatics studies. To enhance the limited generalization ability of typical deep learning frameworks for medical image analysis, unsupervised domain adaptation (UDA) methods have been applied. In this work, we propose to improve the performance of UDA methods on cross-domain neuron membrane segmentation in EM images. First, we designed a feature weight module considering the structural features during adaptation. Second, we introduced a structural feature-based super-resolution approach to alleviating the domain gap by adjusting the cross-domain image resolutions. Third, we proposed an orthogonal decomposition module to facilitate the extraction of domain-invariant features. Extensive experiments on two domain adaptive membrane segmentation applications have indicated the effectiveness of our method. 
\end{abstract}


\section{Introduction}
The segmentation of neuron membrane is essential in neuroanatomical studies. For example, the detection of neuronal membrane is an important step in the reconstruction of neural circuits \cite{ciresan2012deep}. In the task of segmentation of neuronal membrane, the distinction between neuronal membrane and cell nucleus needs to be identified to avoid the segmentation of neuronal nucleus. The deep learning-based approach has made significant progress in cell boundary segmentation tasks \cite{lee2015recursive}. However, high-performance deep learning methods often rely on large amounts of labeled data, and they lack generalization ability on new datasets. These issues are particularly severe in membrane segmentation for electron microscopy (EM) images since there are large distinctions in the EM between different resources, due to the variance in the devices, and the data acquisition processes.

Recently, many unsupervised domain adaptive methods have been proposed for medical image segmentation \cite{mahmood2018unsupervised,dabis,liu2020unsupervised,li2022domain}. Unsupervised domain adaptation aims to accomplish the task on an unlabeled target domain dataset by training on both labeled source domain dataset and unlabeled target domain dataset. Although these methods have been widely employed, they still face the following challenges in cross-domain membrane segmentation in EM images. Firstly, differences in resolution between EM datasets are not considered, which leads to inaccurate feature alignment. Secondly, in the case of small datasets, the under-utilization of geometric features usually leads to a poor prediction of boundary locations \cite{yao2019integrating}, especially in the unsupervised domain adaptation (UDA) tasks where supervised learning for the target images is not available. Finally, they did not consider the effect of the unstable adversarial learning process for the domain discriminator, which might induce the mis-alignment on model performance in domain classification tasks \cite{dabis,liu2022decompose}. Considering the neuron membrane segmentation task and the above challenges, we proposed a structural feature-based domain adaptive approach. 

The major contributions of our work are summarized as follows: \textbf{(1)} Due to the large variation in image quality, we proposed a multi-scale super-resolution network based on structural features for the segmentation task. It can also reduce domain differences by alleviating the resolution differences between different data. \textbf{(2)} We proposed an unsupervised domain-adaptive feature selection strategy. It encourages the model to use a mixture of geometry features on the connection structures for the images. These features are less sensitive to domain bias compared with the raw images, which further facilitates the adaption to assist the segmentation task. \textbf{(3)} For the domain discriminator, we proposed orthogonal decomposition and self-attention methods to reduce the instability in the adversarial learning process and induce the models to generate the domain-invariant features. \textbf{(4)} Our methods have been validated on two cross-domain membrane segmentation settings and outperformed the state-of-the-art methods by a large margin.

\begin{figure*}[htb]
\centering
\includegraphics[width=0.75\textwidth]{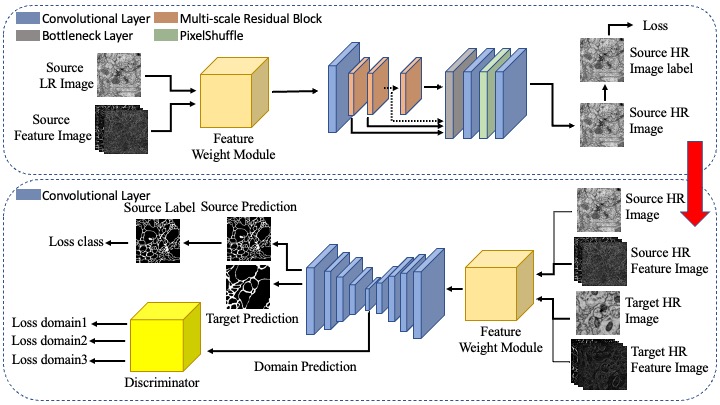}\\
\caption{The framework of our proposed unsupervised domain adaptation method for structural feature-based neuron segmentation.}
\label{frame}
\end{figure*}

\section{Related works}

\subsection{Medical Image Segmentation}

Deep learning methods have received a lot of attention in the field of medical image segmentation due to their powerful representation learning ability. The U-Net method \cite{UNet} has been very successful in medical image segmentation tasks as it can combine low-level features and high-level semantic features very well. Many innovative network architectures have emerged for segmentation based on U-Net, which combine methods such as residuals \cite{diakogiannis2020resunet}, dense linking \cite{denseunet}, self-attention mechanisms \cite{lin2021faster}, and transformer modules \cite{transunet}. Other detection-based methods can segment the biomarker objects at the instance level~\cite{Liu2021PanopticFF,Huang2022LearningTM,He2021CDNetCD}. However, these segmentation methods are not robust on multiple neuronal datasets due to the variability between different datasets.

\subsection{Unsupervised Domain Adaptation}

The goal of domain adaptation is to reduce the distribution gap between the target domain and the source domain and to improve the robustness and generalization ability of the model \cite{mahmood2018unsupervised}, \cite{dann}. Domain adaptation is divided into two main categories: feature alignment and image alignment. Feature alignment learns domain invariant features across domains mainly through CNN models. Most of the models use a domain adversarial neural network (DANN)-like structure \cite{li2019joint}. ADVENT \cite{advent} uses an adversarial learning-based entropy minimization method to reduce the distance between two domains and thus indirectly minimize the entropy value of the target domain. AdaptSeg \cite{adaptseg} makes two segmentation results with approximate distribution by the discriminator. Moreover, adversarial learning approaches are used to jointly train a domain discriminator and a segmenter to learn domain-invariant features for the cross-domain medical image segmentation tasks \cite{dcda}. Image alignment is used for domain adaptation based on image synthesis models~\cite{gan,cycgan,yang2020fda}. The original Generative Adversarial Network (GAN) \cite{gan} uses random noise as input, and the CycleGAN method \cite{cycgan} is proposed, which does not require pairs of training samples. Then, many GAN-based methods for domain-adaptive image generation are further proposed~\cite{mahmood2018unsupervised,liu2020pdam}, which have enhanced the models' generalization ability by utilizing the synthesis images. Additionally, the image synthesis methods based on Fourier transformation have also been widely employed for UDA at the appearance level~\cite{yang2020fda,Yu2022UnsupervisedDA}.

In the domain adaptation task of cell membrane segmentation, DABIS \cite{dabis} uses a DANN-like structure to achieve the segmentation task. Based on DANN, ADACS \cite{haq2020adversarial} uses a decoder network for segmentation prediction results to make the target image and target prediction as correlated as possible.

However, none of these methods takes full advantage of geometric features in the case of small datasets and does not take into account the large differences in resolution between different EM datasets. Such issues are common in cross-domain membrane segmentation in EM images, which limits the applicability of existing UDA methods. Our proposed method takes full use of structural features in the case of small datasets and overcomes the challenge of different resolutions between different EM datasets.

\section{Methodology}



In this section, we present our proposed unsupervised domain adaptation method for neuron segmentation based on structural features. As shown in Fig. \ref{frame}, our method consists of two parts. The first part is a structural feature-based super-resolution image generation network that aligns the images by alleviating the domain gaps at the appearance level. The second part is a structural feature-based image segmentation network for feature alignment.

\subsection{Geometry Feature Enhanced Super-resolution Network}

Since different neuronal datasets are usually generated by different instruments, directly working on the UDA by ignoring the resolution distinctions is harmful since the tubular details might be obscure. Directly aligning features based on these tubular details may ignore domain biases in these specific tubular parts. Therefore, we propose a resolution-based approach to address such a problem. Focusing on improving the resolution of tubular parts can induce accurate feature alignment on the tubular structures. As shown in Fig. \ref{frame}, our super-resolution network is based on a multi-scale residual network \cite{li2018multi}. The inputs to the network are low-resolution images and their multiple corresponding geometry features. The low-resolution images are the ones down-sampled from the original images, and the corresponding multiple feature images are input to the feature weight module. The training process aims at improving the resolution of these low-resolution images, especially tubular components. After training, the model will be directly used on the original images, to get the high-resolution ones with more comprehensive details on the tubular components.

\begin{figure}[t]
\begin{minipage}[b]{1.0\linewidth}
  \centering
  \centerline{\epsfig{figure=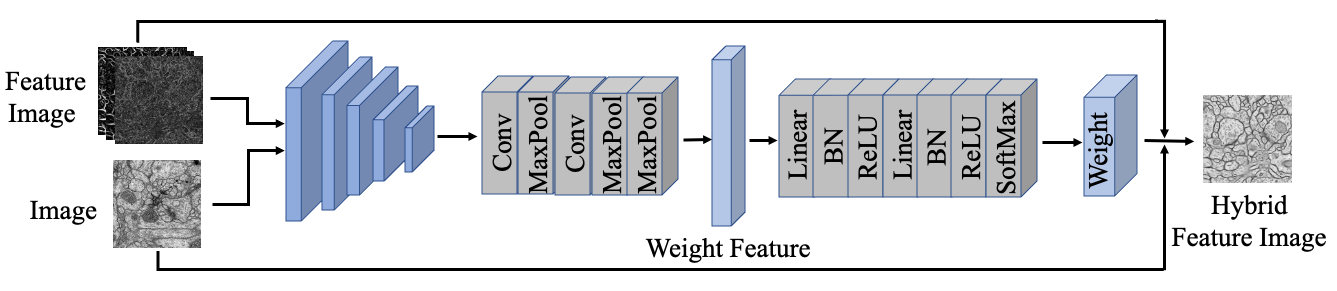,width=\linewidth}}
\end{minipage}
\caption{The architecture of the feature weight module. The input is multiple structural features and the original image. The feature weight module obtains the weight features through multiple convolution blocks and then extracts the weight values by linear blocks. A hybrid feature suitable for the current task and datasets is obtained by overlaying weight-based features.}
\label{feature_module}
\end{figure}

\textbf{Feature Weight Module:} In the membrane segmentation task, we find that the extracellular structure of neurons resembles a tubular structure, so we extract the tubular structural features and edge features to enhance the super-resolution image synthesis. In this paper, we choose to use the Frangi tubular structure feature extraction method \cite{frangi1998multiscale}, Jerman tubular structure feature extraction method \cite{jerman2016enhancement}, and two edge detection operators (Prewitt \cite{prewitt1970object}, Sobel \cite{duda1973pattern}) to extract the structural features. Most of the tubular structure feature extraction methods are based on the Hessian matrix, but different methods have different sensitivity to noise and the delineation of the boundary can be different. Visual examples of different feature extraction methods are shown in Fig. \ref{enhance}. Specifically, the Frangi method achieves the edge detection of the tubular structure by considering the curvature of the tubular structure through the Hessian matrix. The Jerman method is more uniform for the responses within the structure and considers the different contrasts and responses of different parts of the tubular structure. The Prewitt operator suppresses noise in the EM images from various acquisition protocols. The Sobel operator focuses on the different effects of neighboring pixels on each pixel within the tubular structure. A hybrid feature image is defined as below:

\begin{footnotesize} 
\begin{equation}\label{eq1}
\begin{aligned}
F^{H} = \beta \times \left\{ \sum_{i=1}^{n} \left [ \alpha^i \times \left ( 255-F^i\right ) \right] \right\} + \left( 1 - \beta \right) \times F^{Img},
\end{aligned}
\end{equation}
\end{footnotesize}where \( \alpha^i\) is a parameter learned by the feature weight module, representing the weight of the \( i^{th}\) feature of the image, \( \beta\) is the weight of all features, \( n\) is the number of different feature extractors, and \( F^{H} \) is a hybrid feature image. We propose feature weight modules to incorporate the different characteristics of these features for a comprehensive hybrid feature. Fig. \ref{feature_module} shows the structure of the feature weighting module.

In the feature weight module, the size of the feature map obtained after multiple convolutions and pooling of multiple images is \( b \times 512\times h \times w\), and then the feature map is resized to \( b \times 2048 \times 1 \times 1\) by Convolution and Max Pooling. \( b, h, w\) denote the batch size, height, and width, respectively. Finally, the weight values of each feature are obtained by linear transformation and activation function. The weight values are superimposed on the corresponding features and fused with the original images to obtain the hybrid feature image (HFI). Since the pixel values of the tubular part in the original image are close to \(0\), and those of the tubular part in the structural feature are close to \(255\), the structural features should be inverted.

\textbf{Image Super-resolution Network:} After acquiring the hybrid features, we use a super-resolution network to generate synthesis images with detailed tubular structures which are robust to domain differences. Our super-resolution network is based on a multi-scale residual network \cite{li2018multi}. The upper part of Fig. \ref{frame} is a multi-scale residual super-resolution network based on structural features. The input of the network is a low-resolution image generated by downsampling and cubic interpolation of the original image and the corresponding multiple features. The output of the super-resolution network is a high-resolution generated image (HRG). Firstly, a hybrid feature is obtained by the feature weight module. Then the feature maps of eight different scales are obtained by eight sets of multi-scale Residual block \cite{li2018multi} after convolutional layer. After putting the eight different scales of feature maps into the bottleneck layer, the convolution operation and PixelShuffle are performed. The label for training the super-resolution network is defined as below:

\begin{equation}\label{eq2}
\begin{aligned}
S^{HR\_Label}=\eta \times S^{HR}+\left( 1 - \eta \right) \times S^{Seg},
\end{aligned}
\end{equation}
where \( \eta \) is the weight parameter, \( S^{HR} \) represents high-resolution image, and \( S^{Seg} \) represents the segmentation label. We use the L1-loss as the supervised loss for learning the super-resolution target. By enhancing the weight of the tubular part for both the input and the label, the entire super-resolution network focuses on improving the clarity of the tubular part.

\begin{figure}[t]
\begin{minipage}[b]{1.0\linewidth}
  \centering
  \centerline{\epsfig{figure=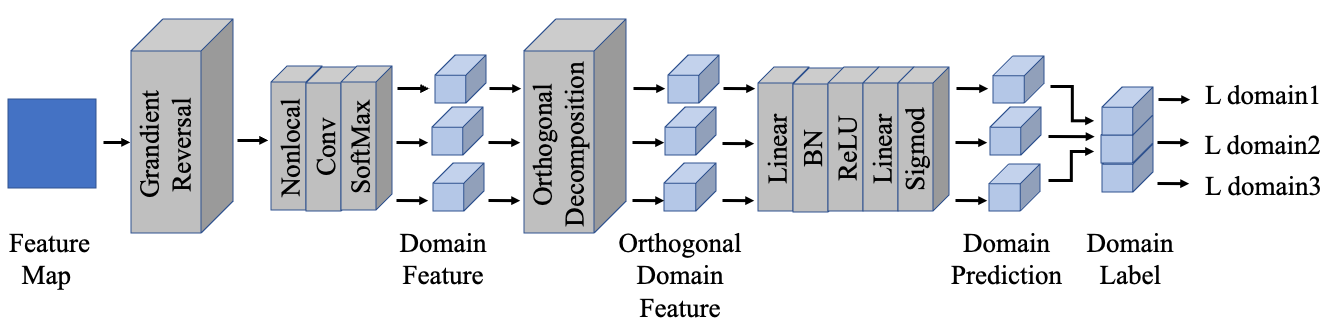,width=\linewidth}}
\end{minipage}
\caption{The detailed structure of the discriminator.}
\label{Optimization}
\end{figure}

\subsection{Segmentation Networks based on Feature Weight Modules and High-resolution Generated Images}

Feature images as a prior knowledge are effective in domain adaptation segmentation tasks based on the small amount of medical image data. All images from the source and target domains are passed through the geometry feature-enhanced super-resolution network in Section III-A to obtain geometry feature-enhanced high-resolution generated images. All feature extraction methods mentioned in Section III-A are applied to the high-resolution generated images to obtain the corresponding feature images. As shown in the bottom part of Fig. \ref{frame}, our segmentation network is based on U-Net, and the input to the network is high-resolution generated images and multiple feature images. In the network, multiple features need to be first transformed into a hybrid feature by the feature weight module. This feature weight module has the same structure as defined in Section III-A, and the parameters are not shared. In addition, the self-attention and orthogonal decomposition mechanism are proposed in the domain adversarial module to achieve feature alignment.

In the field of domain adaptation based on adversarial learning, the basic domain classification module consisting of multi-linear, normalization function and activation function will ignore some acquisition and judgment of context information. Due to the unstable learning process, the decision of the adversarial domain classifier is inaccurate and the direction of the gradient is not always optimal \cite{arjovsky2017wasserstein}. In response to the above problems, we use the self-attention mechanism \cite{nonlocal} to help the domain classifier to better obtain information about long-range dependencies. To stabilize the adversarial learning process and achieve accurate feature adaption, the orthogonal-based decomposition modules are further proposed for domain classifier optimization (DCO).

Fig. \ref{Optimization} shows the detailed structure of the domain classifier. First, we pass the feature map of the U-Net bottom layer to the Non-local module \cite{nonlocal}, so that the feature map can take into account the long-range dependencies information that contributes to the filtered response. Then we get three vectors from the feature map, and Schmidt orthogonalization is performed on these three vectors. These features will pass through the linear layer, the batch normalization layer, and the activation function to finally get three domain classification losses. The loss computations in the segmentation network are defined as follows:

\begin{align}\label{eq4}
L_s &= L_s^{Seg}+GRL\left(\mu_1L_s^{D1}+\mu_2L_s^{D2}+\mu_3L_s^{D3}\right)\\
L_t &= GRL\left(\mu_4L_t^{D1}+\mu_5L_t^{D2}+\mu_6L_t^{D3}\right)\\
L &= L_s+L_t.
\end{align}

All losses are based on binary cross-entropy loss. The total loss (\(L\)) consists of source loss (\(L_s\)) and target loss (\(L_t\)). \(L_s\) is the loss of the source, which consists of one segmentation loss (\(L_s^{Seg}\)) and three domain losses (\(L_s^{D1}\), \(L_s^{D2}\) and \(L_s^{D3}\)). GRL is the gradient reversal layer. \(L_t\) is the target loss, which is composed of three domain losses (\(L_t^{D1}\), \(L_t^{D2}\) and \(L_t^{D3}\)). \(\mu_n\) are controllable parameters.

\section{EXPERIMENT AND RESULTS}

\subsection{Datasets}

To fully validate the effectiveness of the proposed method, we tested our method and comparison methods on two neuronal membrane datasets: ISBI 2012 EM Segmentation Challenge (ISBI) \cite{ciresan2012deep} and Mouse Piriform Cortex EM datasets (Piriform) \cite{lee2015recursive}. The ISBI dataset has one 3D volume with 30 slices, and we used 2D slices for training and testing, with 20 for training and 10 for testing. The size of 2D slices of the ISBI dataset is \( 512 \times 512 \). There are 4 cases of 3D data in the Piriform dataset (167 slices, 169 slices, 169 slices, and 120 slices, respectively). Because of the high similarity of adjacent slices, we extracted one 2D slice from every five slices in the 3D volume, of which 100 cases were used for training and 20 cases for testing. There are 2 sizes of 2D slices of the Piriform dataset, \( 256 \times 256 \) and \( 512 \times 512 \).

\subsection{Implementation Details}

Due to the small amount of ISBI data and the unbalanced number of samples in the two datasets, we increased the ISBI training dataset from 20 to 100 using rotation and flipping. The input size of both fields in a cross-domain study should be the same, so we change the data size to \( 512 \times 512 \) using cubic interpolation for all the data. We implemented our method using PyTorch. We used the Adam optimization method with an initial learning rate of 0.01 and a weight decay rate of 0.001. \( \beta \) is set to 0.5 and \( \eta \) is set to 0.9. The values of \( \mu_1 \) to \( \mu_6 \) are set to 0.03. The Dice similarity coefficient (Dice) is employed to evaluate the segmentation performance. In addition, we also use the pixel-wise Hausdorff\_95 distance (95HD) as our evaluation metrics. 

\subsection{Comparison Experiments}

\begin{figure*}[htb]
\centering
\centerline{\epsfig{figure=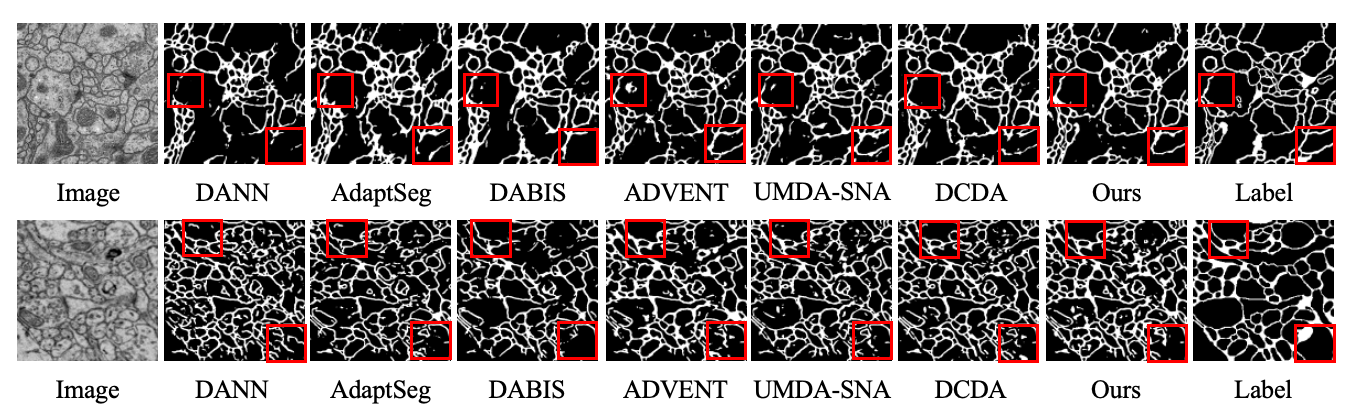,width=0.9\linewidth}}
\caption{Visualization examples of comparative experiments. Top row: the results under the Pirform $\rightarrow$ ISBI setting. Bottom row: the results under the ISBI $\rightarrow$ Piriform setting.}
\label{duibi}
\end{figure*}

To validate the superiority of the proposed method, we conducted a comprehensive comparison with five unsupervised domain adaptive methods on two datasets, including DANN \cite{dann}, AdaptSeg \cite{adaptseg}, DABIS \cite{dabis}, ADVENT \cite{advent}, UMDA-SNA \cite{umda-sna}, DCDA \cite{dcda}. We use the same U-Net backbone in these methods for a fair comparison. We show the upper and lower bounds of the metrics for U-Net without domain adaptation (without-domain-adaptation (Source Only) and fully supervised (Oracle)). Our proposed framework achieves a considerable performance improvement (Dice from 62\% to 74\% and from 49\% to 57\%). Our approach outperforms the mainstream unsupervised domain adaptation methods in terms of metrics. The generated images reduce the domain differences while facilitating the models to learn information about the tubular structure. Fig. \ref{duibi} shows visualization examples of comparative experiments with different unsupervised domain adaptation methods on two neuronal membrane datasets. Table \ref{tab_comparison} quantitatively shows the difference in metrics of different methods. It illustrates that our method produces more accurate segmentation results from the qualitative level.

\subsection{Ablation Studies}

\begin{table}[t]
\begin{center}
\caption{Comparison Experiments with other UDA semantic segmentation methods.} \label{tab_comparison}
\resizebox{1.0\linewidth}{!}{
\begin{tabular}{|c|cc|cc|}
\hline
\multirow{2}{*}{Method} & \multicolumn{2}{c|}{Piriform $\rightarrow$ ISBI}                        & \multicolumn{2}{c|}{ISBI $\rightarrow$ Piriform}                     \\ \cline{2-5} 
                        & Dice~$\uparrow$      & 95HD~$\downarrow$   & Dice~$\uparrow$     & 95HD~$\downarrow$   \\ \hline
Source Only             & 62.68          & 12.89          & 49.87          & 13.41          \\ \hline \cline{1-5}
DANN \cite{dann}        & 65.53          & 12.37          & 51.01          & 13.09          \\ \hline
AdaptSeg \cite{adaptseg}& 67.63          & 12.30          & 52.82          & 13.04          \\ \hline
DABIS \cite{dabis}      & 69.83          & 12.08          & 53.56          & 12.77          \\ \hline
ADVENT \cite{advent}    & 70.54          & 11.86          & 54.79          & 12.63          \\ \hline
UMDA-SNA \cite{umda-sna}& 72.46          & 12.02          & 55.39          & 12.75          \\ \hline
DCDA \cite{dcda}        & 73.73          & 11.49          & 56.31          & 12.60          \\ \hline
Ours                    & \textbf{74.67} & \textbf{11.32} & \textbf{57.47} & \textbf{12.30} \\\cline{1-5} \cline{1-5}
Oracle                  & 76.32          & 10.54          & 64.50          & 11.54          \\ \hline
\end{tabular}}
\end{center}
\end{table}

\begin{figure*}[ht]
\centering
\centerline{\epsfig{figure=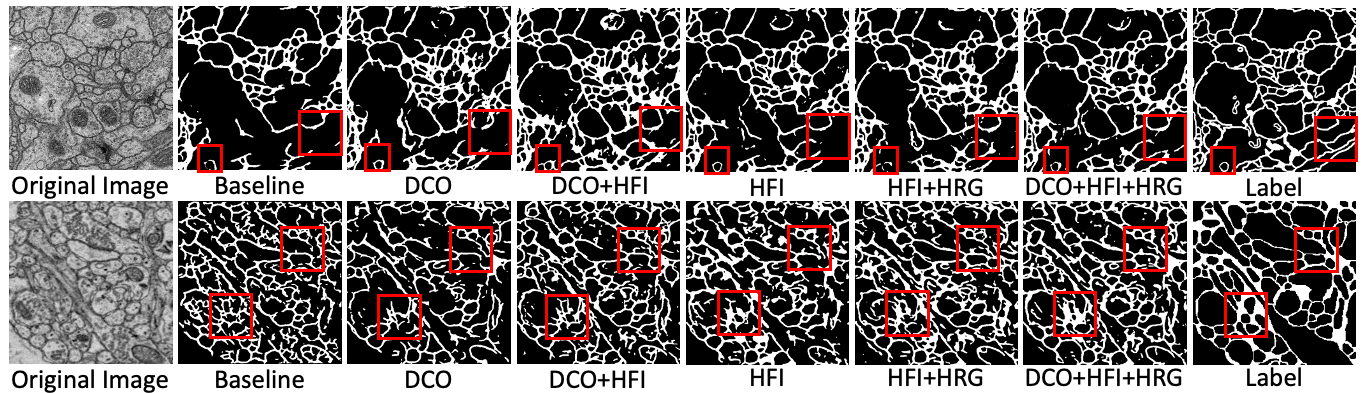,width=0.85\linewidth}}
\caption{Visualization examples of ablation studies. The top row is the results under the Pirform $\rightarrow$ ISBI setting. The bottom row is the results under the ISBI $\rightarrow$ Piriform setting.}
\label{xiaorong}
\end{figure*}

\begin{table}[t]
\begin{center}
\caption{Ablation studies for our proposed DCO, HFI, and HRG modules} \label{tab_ablation}
\resizebox{\linewidth}{!}{
\begin{tabular}{|c|cc|cc|}
\hline
\multirow{2}{*}{Method} & \multicolumn{2}{c|}{Pirform $\rightarrow$ ISBI} & \multicolumn{2}{c|}{ISBI $\rightarrow$ Piriform}       \\ \cline{2-5} 
                        & Dice~$\uparrow$        & 95HD~$\downarrow$  & Dice~$\uparrow$     & 95HD~$\downarrow$    \\ \hline
Baseline                & 65.53          & 12.37          & 51.01          & 13.09          \\ \hline
DCO                     & 66.66          & 12.02          & 51.64          & 13.00          \\ \hline
DCO+HFI                 & 72.46          & 11.95          & 56.36          & 12.75          \\ \hline
HFI                     & 71.60          & 12.08          & 55.60          & 12.77          \\ \hline
HFI+HRG                 & 73.50          & 11.88          & 57.31          & 12.65          \\ \hline
DCO+HFI+HRG             & \textbf{74.67} & \textbf{11.32} & \textbf{57.47} & \textbf{12.30} \\ \hline
\end{tabular}}
\end{center}
\end{table}

\begin{figure}[t]
\begin{minipage}[b]{1.0\linewidth}
  \centering
  \centerline{\epsfig{figure=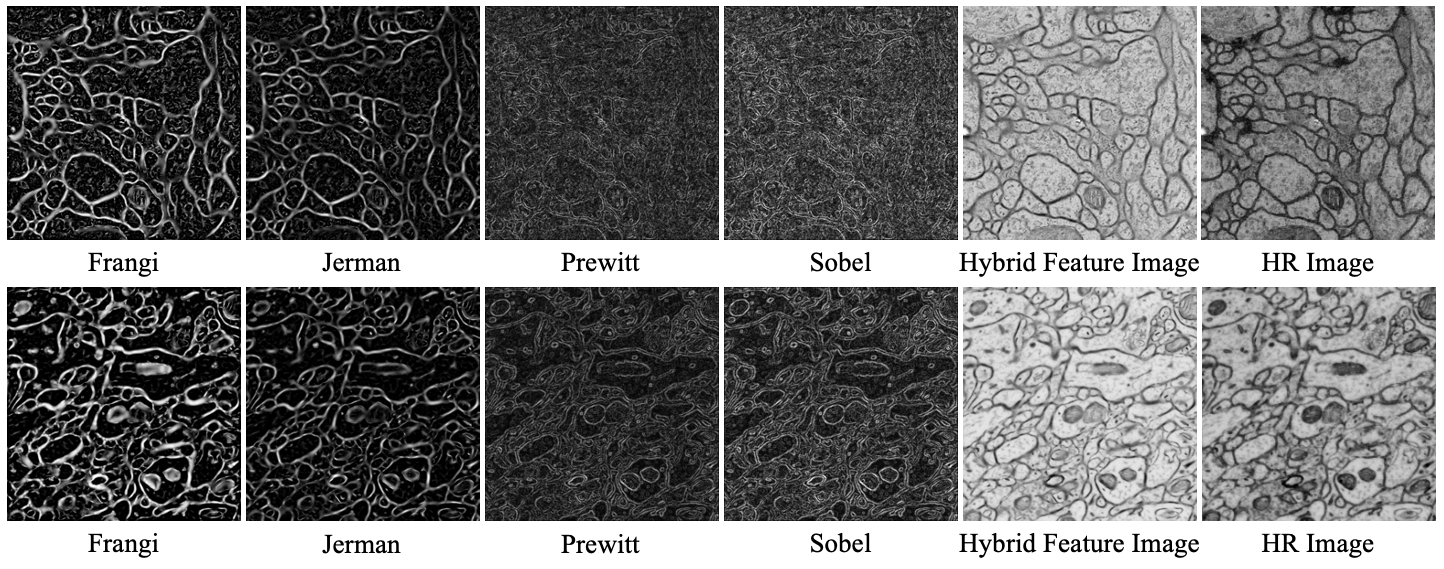,width=\linewidth}}
\end{minipage}
\caption{Visualization results of different structural feature extraction methods and results of structural feature-based super-resolution network.}
\label{enhance}
\end{figure}

To demonstrate the effectiveness of our method, we conduct ablation experiments on the Piriform $\rightarrow$ ISBI and ISBI $\rightarrow$ Piriform. As shown in Table \ref{tab_ablation}, we use U-Net-based DANN as the backbone, and we attach the proposed sequence of modules to the backbone network. It shows the improvement of segmentation accuracy by domain classifier optimization (DCO), Hybrid Feature Image (HFI), and High-resolution Generated Image (HRG) of our method. In addition, it can be found that all the Dice and 95HD performance has been effectively improved by introducing our proposed modules, and finally the best performance is obtained by jointly employing all proposed modules. Fig. \ref{xiaorong} presents the visualization results of the ablation experiment. Fig. \ref{enhance} shows the visualization examples of the feature weight module and the results of the super-revolution network.

\section{Conclusion}

Cross-domain neuronal membrane segmentation in EM images is crucial in neuroanatomical studies. In this paper, we proposed an unsupervised domain adaptation method based on DCO, HFI, and HRG. The method implements domain adaptation from both image alignment and feature alignment perspectives. To the best of our knowledge, we are making an early attempt to use tubular structural features and edge features to assist super-resolution networks to clarify specific parts and reduce domain differences. We designed the feature weight module to integrate the geometry characteristic from different features. In order to better solve the segmentation task of domain adaptation, high-resolution generated images and hybrid features are used. Experiments on the two cross-domain membrane segmentation settings indicate the effectiveness of our method for the unsupervised domain adaptation task of neuronal membrane segmentation. Given the appealing performance of our method, it can be extended to other cross-domain tubular analysis tasks for medical image analysis in future works.

\bibliographystyle{IEEEtran}
\bibliography{IEEEexample}

\end{document}